\documentclass[letterpaper, 10 pt, conference]{ieeeconf}  

\IEEEoverridecommandlockouts                              

\usepackage{graphics} 
\usepackage{epsfig} 
\usepackage{amsmath} 
\usepackage{amssymb}  
\usepackage{xcolor}
\usepackage{booktabs}
\usepackage{subcaption}
\usepackage{multirow}

\title{\LARGE \bf
ExoLaN: Physics-Consistent Context-Aware\\Dynamics Learning for Exoskeletons
}

\author{
  Lucas Schulze$^{1,2}$,
  Maximilian Schwarz$^{1*}$,
  Jona Hoppe$^{1*}$, 
  Jan Peters$^{1,2,3,4}$,
  Oleg Arenz$^{1,2}$
\thanks{
$^{*}$Equal contribution.
\newline This work was funded by the German Research Foundation (DFG) - Project number PE 2315/18-1, and by the German Federal Ministry of Research, Technology and Space (BMFTR) - Project number 01IS23057B, and as part of the Robotics Institute Germany (RIG). This project has been supported by a hardware donation by NVIDIA through the Academic Grant Program.
\newline$^{1}$Department of Computer Science, Technical University of Darmstadt, Germany.
$^{2}$Robotics Institute Germany (RIG).
$^{3}$Hessian.AI.
$^{4}$German Research Center for AI (DFKI), Research Department: Systems AI for Robot Learning.
\newline Corresponding author: {\tt\small lucas.schulze@robot-learning.de}
}
}

\newcommand{\Transp}{^\mathrm{\scriptscriptstyle T}}
\newcommand{\Real}{\mathbb{R}}
\newcommand{\NegSciExp}[1]{\mathrm{e}{\text{-}#1}}
\newcommand{\PosSciExp}[1]{\mathrm{e}{#1}}
\newcommand{\NegSciExpB}[1]{\boldsymbol{\mathrm{e}}\textbf{-}#1}

\newcommand{\jointTau}{\boldsymbol{\tau}}
\newcommand{\jointPos}{\mathbf{q}}
\newcommand{\jointVel}{\dot{\mathbf{q}}}
\newcommand{\jointAcc}{\ddot{\mathbf{q}}}

\newcommand{\jointTauHuman}{\boldsymbol{\tau}_\mathrm{h}}
\newcommand{\jointTauExo}{\boldsymbol{\tau}_\mathrm{e}}
\newcommand{\jointTauContact}{\boldsymbol{\tau}_\mathrm{c}}

\newcommand{\genTau}{\jointTau}
\newcommand{\genPos}{\jointPos}
\newcommand{\genVel}{\jointVel}
\newcommand{\genAcc}{\ddot{\mathbf{q}}}

\newcommand{\PosHip}{q_\mathrm{hip}}
\newcommand{\PosKnee}{q_\mathrm{knee}}

\newcommand{\JacContact}{\mathbf{J}_\mathrm{c}}
\newcommand{\ForceContact}{\mathbf{f}_\mathrm{c}}

\newcommand{\EoMC}{\mathbf{c}}
\newcommand{\EoMG}{\mathbf{g}}
\newcommand{\Lag}{\mathcal{L}}
\newcommand{\EKin}{K}
\newcommand{\EPot}{P}

\newcommand{\inertiaMat}{\mathbf{H}}
\newcommand{\inertiaTri}{\mathbf{L}}

\newcommand{\paramInertia}{\boldsymbol{\theta}_\inertiaMat}
\newcommand{\paramPot}{\boldsymbol{\theta}_\EPot}
\newcommand{\paramLstm}{\boldsymbol{\theta}_\mathbf{z}}
\newcommand{\paramInsole}{\boldsymbol{\theta}_\mathbf{I}}
\newcommand{\paramAll}{\boldsymbol{\theta}}
\newcommand{\FID}{f_\mathrm{ID}}
\newcommand{\FFD}{f_\mathrm{FD}}

\newcommand{\latentVariableZ}{\mathbf{z}}

\newcommand{\ForceInsole}{f_v}
\newcommand{\CopInsole}{\mathbf{c}_\mathrm{xy}}
\newcommand{\InsoleLearnedMap}{\mathbf{B}}

\newcommand{\WeightFD}{w_\mathrm{fd}}
\newcommand{\WeightMS}{w_\mathrm{ms}}

\newcommand{\Subsamples}{N_s}
\newcommand{\MShorizon}{H}
\newcommand{\NormTorque}{W_m}
\newcommand{\NormQp}{p}
\newcommand{\NormQv}{v}

\begin{document}

\maketitle
\thispagestyle{empty}
\pagestyle{empty}

\begin{abstract}
Task-agnostic assistive exoskeleton control based on human intention offers greater flexibility than conventional approaches that rely on predefined tasks or motion patterns. Human joint torque estimation enables task-agnostic assistance by characterizing user actions. Physics-consistent methods such as Deep Lagrangian Networks (DeLaN) have been applied to estimate the human torques in multi-user settings, but existing approaches cannot adapt to a specific user without retraining, and do not account for intermittent contacts during locomotion. We propose ExoLaN, a Context-Aware DeLaN for human-exoskeleton interaction that learns the full coupled system dynamics while adapting to changes in interaction context. ExoLaN combines temporal context with partial contact-force measurements from force-sensitive insoles to infer latent dynamics embeddings and estimate generalized contact torques. On seven unseen users performing 21 unseen tasks, ExoLaN reduces torque estimation MSE by 7\% compared to a black-box baseline. Beyond inverse dynamics, ExoLaN serves as a unified model that also enables accurate forward prediction: training with a multi-step prediction loss reduces acceleration MSE by 59\% and long-horizon position and velocity errors by 60\% and 93\%, respectively, compared with a single-step loss. Moreover, the learned latent context captures task information without explicit task labels, making it a promising signal for task-aware assistive control.
\end{abstract}


\section{INTRODUCTION}
Lower-limb exoskeletons can augment mobility, reduce physical effort, and support rehabilitation, making them important assistive tools in medicine and industry~\cite{Siviy2022}. 
Conventional approaches for assistive control rely on predefined and structured tasks, requiring task-specific control modules and switching mechanisms between them~\cite{event_driven_gait_assist}, as well as user-specific tuning~\cite{sit-to-stand}.
A task-agnostic alternative is to estimate the human's intended torques, and use this estimate to provide assistance toward the intended action.

Human torque can be estimated using first-principles models of the coupled human-exoskeleton system~\cite{brubaker_2007, sarmiento_2024, forward_dyn_sarajchi2025}. These approaches typically rely on rigid-body dynamics and require assumptions about the system's geometry, inertial parameters, coordinate definitions, and sensor configuration. Moreover, the flexible nature of human muscular actuation and the compliance of both the human body and exoskeleton make accurate modeling of human-exoskeleton challenging~\cite{Firouzi2025}, often requiring subject-specific parameter identification or calibration.

Data-driven approaches offer greater flexibility in system representation, without requiring prior knowledge of the underlying dynamics, and can leverage multimodal sensor information~\cite{lstm_prediction, tcn_only_torque_estimation, actor-critic_prediction, acceleration_prediction}. However, black-box models can exhibit limited generalization to users and tasks outside the training distribution, while providing limited physical interpretability, which is particularly relevant for safe and predictable human-robot interaction~\cite{safety_hri_survey}.
\begin{figure}[t]
    \centering
    \includegraphics{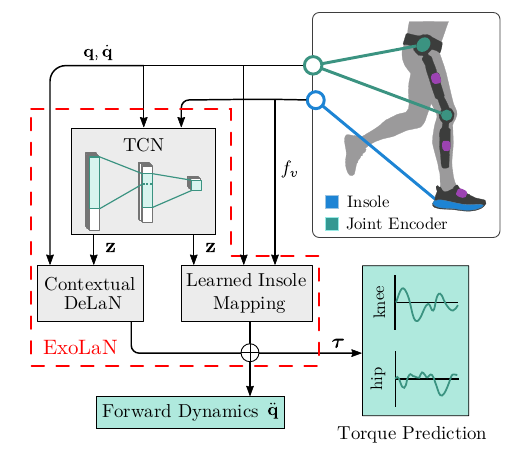}
    \caption{Overview of ExoLaN for human-exoskeleton dynamics modeling. Contextual DeLaN denotes the context-conditioned DeLaN proposed in \cite{cadelac}. The exoskeleton diagram is adapted from \cite{tcn_nature_2024}.}
    \label{fig:exolan_architecture}
\end{figure}

Grey-box methods such as Deep Lagrangian Networks (DeLaN)~\cite{delan_lutter2018} provide an alternative by embedding physical structure into data-driven dynamical models. This allows DeLaN to learn a local approximation of complex system dynamics directly from data while preserving physical properties, such as energy conservation, as also demonstrated in soft robotic systems~\cite{soft_robots}.


For exoskeletons, \cite{delan_exoskeleton_single_model} learns DeLaN for a single exoskeleton from data collected without a wearer, then combines the learned model with residual Gaussian process regression for motion tracking using a backstepping controller. Thus, the learned dynamics characterize the exoskeleton itself, while human-exoskeleton interaction is treated as an external effect during runtime.

In a multi-user setting, \cite{pilan} proposes PiLaN based on DeLaN, which uses a finite-difference approximation of the Lagrangian to estimate joint accelerations without requiring torque measurements. A single model is trained using data from ten users and evaluated on eight new users across both in-distribution and unseen tasks.
The authors reported improved motion intention estimation and tracking performance compared with black-box baselines, demonstrating generalization across new users and tasks.

In robotics, recent work \cite{cadelac} proposed Context-Aware DeLaN, which extends DeLaN to environments with changing dynamics by conditioning the dynamics model on a latent context representation. A recurrent context encoder performs online system identification to infer this context from history, allowing a single DeLaN model to represent different systems. The approach further leverages a nominal robot model to learn only the residual dynamics associated with changing loads. The resulting context-aware model is integrated with model predictive control (MPC) to enable adaptive trajectory tracking under varying loads in manipulation tasks.

While Context-Aware DeLaN enables adaptation to changing system dynamics, its application to robotic systems assumes a nominal model and focuses on residual dynamics arising from changing loads. This setting does not directly address the user-dependent dynamics of a coupled human-exoskeleton system or the intermittent contacts encountered during human locomotion. Moreover, existing exoskeleton applications do not fully exploit the ability of a single DeLaN model to provide both inverse and forward dynamics, which could enable torque estimation from observed motion as well as future-state prediction.




Therefore, in this work, we introduce ExoLaN, a Context-Aware DeLaN for human-exoskeleton. Unlike \cite{cadelac}, which exploits a nominal robot model to learn residual dynamics associated with changing payloads, ExoLaN learns the full dynamics of the coupled human-exoskeleton system. To account for intermittent contacts during locomotion, we learn a mapping from partial contact force measurements to the corresponding generalized contact torques. The inferred context conditions the learned dynamics, enabling the model to adapt to changes in human-exoskeleton dynamics across users and task conditions.

We train ExoLaN on a multi-task exoskeleton dataset \cite{tcn_nature_2024} and evaluate its performance across seven held-out participants on activities presented during training and 21 excluded activities. We compare ExoLaN to black-box baselines to evaluate the benefits of having physical structure in the model, examine different context encoders, and the contribution of partial contact force mapping.
%
We further evaluate the model for forward dynamics prediction and investigate the effect of multi-step prediction loss on long-horizon rollouts. The results show that temporal context and measured contact information provide the most consistent improvements for human torque estimation, while the physical structure of DeLaN improves generalization to unseen tasks. Finally, we demonstrate that the learned latent representation contains information about the performed task, highlighting its potential for task-adaptive assistive control.


\subsection{Contributions}
The main contributions of this work are:
\begin{itemize}
    \item We introduce ExoLaN, a Context-Aware DeLaN for human-exoskeleton that uses a temporal convolutional network (TCN) to capture temporal context and estimate human joint torques across users and tasks.
    \item A Learned Insole Mapping (LIM) is introduced to map partial contact force measurements to the corresponding generalized contact torques.
    \item We show the benefits of using a single trained DeLaN for both inverse and forward dynamics, and show that a multi-step prediction loss substantially improves long-horizon prediction.
    \item We demonstrate that the learned latent representation contains task-related information that can be exploited as a signal for task-adaptive assistive control.
\end{itemize}

The remainder of this paper is organized as follows. Section~\ref{sec:lower_lib} presents the considered 2-DoF lower-lib exoskeleton and its dynamical model through Euler-Lagrange. 
Section~\ref{sec:context_aware_exoskeleton} briefly reviews DeLaN, Context-Aware DeLaN, and introduces ExoLaN, including the formulation for learning the full human-exoskeleton dynamics and accounting for contact forces. Section~\ref{sec:experiments} describes the experimental setup and discusses the results. Finally, Section~\ref{sec:conclusion} summarizes our findings and discusses future work.

\section{Lower-limb Exoskeleton}
\label{sec:lower_lib}

\begin{figure}[ht]
    \centering
    \includegraphics[width=0.4\textwidth]{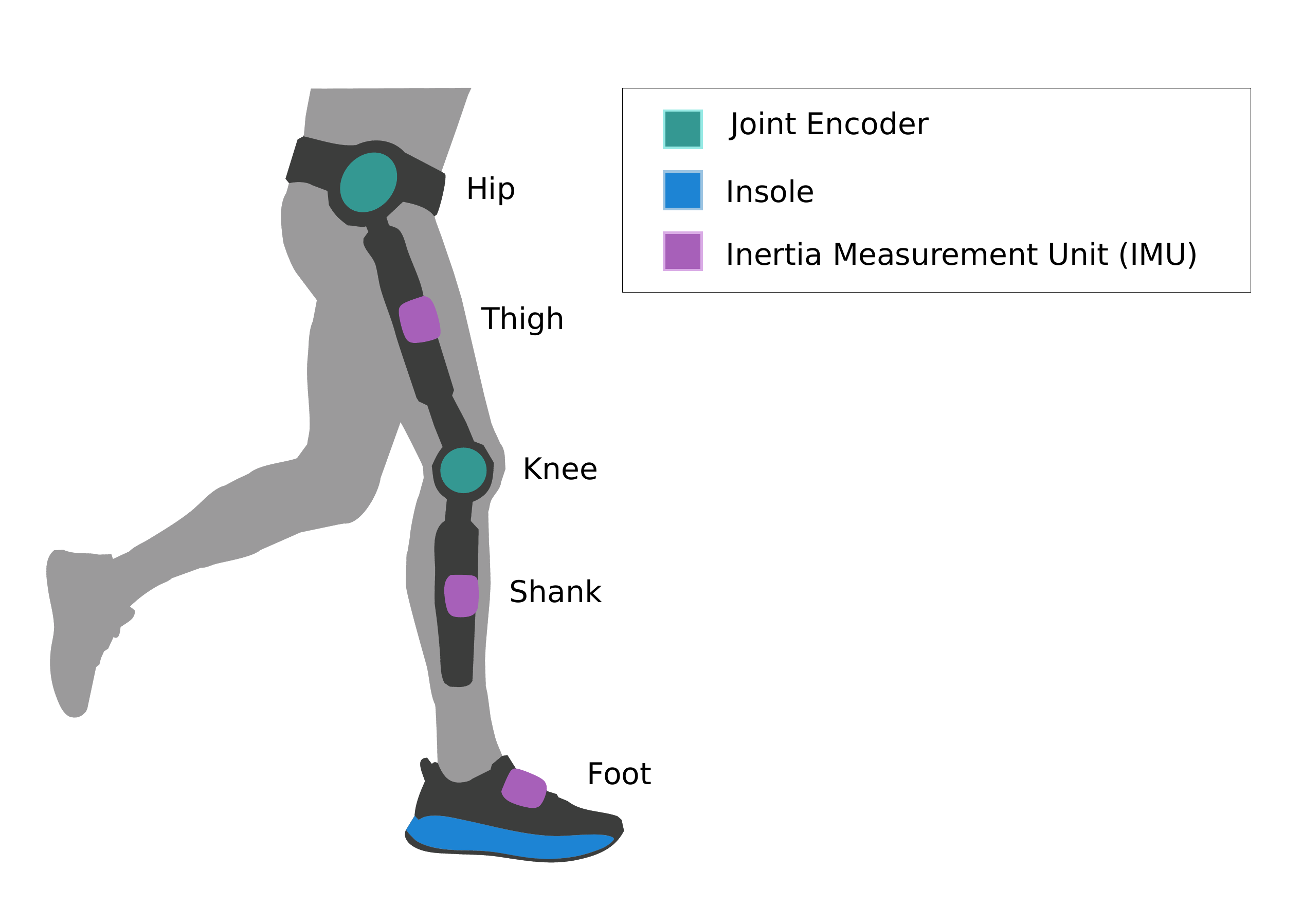}
    \caption{Considered 2-DoF Lower-limb exoskeleton. Diagram adapted from~\cite{tcn_nature_2024}.}
    \label{fig:exoskeleton}
\end{figure}

In this work, we consider the 2-DoF lower-limb powered exoskeleton presented in \cite{tcn_nature_2024}, depicted in Fig.~\ref{fig:exoskeleton}. The exoskeleton has a semi-rigid structure and supports a wide range of cyclic and non-cyclic human motor tasks. It is equipped with joint encoders at the hip and knee joints, six inertial measurement units (IMUs) distributed across the thighs, shanks, and feet, and a pair of wireless force-sensitive insoles. Four quasi-direct-drive actuators provide assistance at the hip and knee joints, with up to 15Nm of sagittal-plane torque per joint.

\subsection{Dynamical model}
The physical coupling between the human and exoskeleton defines a single mechanical system. We model each leg as an independent system, whose generalized coordinates are defined as the hip and knee joint positions, i.e., $\genPos = [\PosHip, \PosKnee]\Transp$. This simplification is motivated by the fact that a whole-body formulation would require additional sensing to estimate the pose and motion of the human pelvis. 

To describe the dynamics of the coupled system, we formulate its Lagrangian $\Lag$ as the difference between the kinetic $\EKin = \frac{1}{2}\jointVel\Transp\inertiaMat(\jointPos)\jointVel$ and the potential $\EPot$ energies. The equations of motion can then be obtained using the Euler-Lagrange equation,
\begin{equation}
    \label{eq:eom_lag_derivative}
    \frac{\mathrm{d}}{\mathrm{d}t} \frac{\partial \Lag(\genPos, \genVel)}{\partial\genVel} -\frac{\partial \Lag(\genPos, \genVel)}{\partial\genPos} = \genTau\text{,}
\end{equation}
where $\inertiaMat$ is the inertia matrix, and $\genTau$ denotes the generalized forces acting on the system. 

For the coupled human-exoskeleton system, the generalized forces can be decomposed into human torques $\jointTauHuman$, exoskeleton torques $\jointTauExo$, and contact torques $\jointTauContact$ arising from intermittent ground contact. Equation~\eqref{eq:eom_lag_derivative} can therefore be written in the form
\begin{equation}
    \label{eq:eom_exo_decomposed}
    \inertiaMat(\jointPos) \jointAcc + \EoMC(\jointPos,\jointVel) + \EoMG(\jointPos)= \genTau = \jointTauHuman + \jointTauExo + \jointTauContact\text{,}
\end{equation}
where $\EoMC(\jointPos,\jointVel)$ represents the Coriolis and centrifugal forces, and $\EoMG(\jointPos)$ represents the gravitational forces.




\section{Context-Aware DeLaN for Exoskeleton}
\label{sec:context_aware_exoskeleton}

In this section, we briefly review DeLaN~\cite{delan_lutter2018,delan_2021} and Context-Aware DeLaN~\cite{cadelac}, and then introduce ExoLaN, our extension for learning the full dynamics of the coupled human-exoskeleton system under intermittent contact.

\subsection{Deep Lagrangian Networks}
To learn physically consistent dynamical models, DeLaN embeds Lagrangian mechanics into a deep learning framework. 
DeLaN estimates $\inertiaMat(\jointPos)$ and $\EPot(\jointPos)$ by two neural networks with learnable parameters $\paramInertia$ and $\paramPot$, respectively.

To guarantee that $\inertiaMat$ is positive definite, the inertia matrix is parameterized through its Cholesky factor $\inertiaTri$ as
\begin{equation}
    {\inertiaMat}(\jointPos, \paramInertia) = \inertiaTri(\jointPos,\paramInertia){\inertiaTri}(\jointPos,\paramInertia)\Transp\text{,}
\end{equation}
where a softplus activation function and an offset are applied to the diagonal elements of ${\inertiaTri}$. The potential energy ${\EPot}(\jointPos, \paramPot)$ is estimated directly by the output of the second network.

Given the learned inertia and potential energy functions, automatic differentiation can be applied to the learned Lagrangian $\Lag$ to obtain a learned equation of motion \eqref{eq:eom_exo_decomposed} through the Euler-Lagrange~\eqref{eq:eom_lag_derivative}. Thus, a single trained DeLaN model provides both the inverse and forward dynamics functions,
\begin{equation}
    \label{eq:delan_inv_dyn_function}
    \genTau = \FID(\genPos, \genVel, \genAcc, \paramInertia, \paramPot),
\end{equation}
\begin{equation}
    \label{eq:delan_fwd_dyn_function}
    \genAcc = \FFD(\genPos, \genVel, \genTau, \paramInertia, \paramPot).
\end{equation}

\subsection{Context-Aware DeLaN}

A DeLaN model represents one mechanical system through its learned Lagrangian $\Lag$, and therefore cannot explicitly account for variations in the underlying system dynamics. This limitation can be addressed by considering a family of mechanical systems described by a parameterized Lagrangian, where a latent variable $\latentVariableZ$ determines the specific dynamics of the system. If $\latentVariableZ$ remains constant over a given time horizon, the system can be regarded as time-invariant, i.e., $\dot{\latentVariableZ}=0$. Consequently, the $\latentVariableZ$ can be incorporated into $\Lag$ without changing the Euler-Lagrange formulation in \eqref{eq:eom_lag_derivative}.

Based on this formulation, \cite{cadelac} proposes Context-Aware DeLaN, which consists of a Contextual DeLaN and a context encoder. The Contextual DeLaN extends DeLaN by conditioning the learned dynamics on the latent context variable $\latentVariableZ$, allowing a single model to represent multiple mechanical systems. The context encoder estimates $\latentVariableZ$ with a long short-term memory (LSTM)~\cite{hochreiter1997long} that takes a historical sequence of system states and residual torques obtained from a nominal model. The resulting model is integrated into an MPC for adaptive robot manipulation.



In this work, we compare the use of TCN~\cite{original_tcn} as a context encoder, due to its reported improved performance over recurrent networks, such as LSTM.




\subsection{Learning Full Dynamics}

The Context-Aware DeLaN proposed in \cite{cadelac} addresses model mismatch due to unknown payloads in robot manipulation tasks. To estimate the context $\latentVariableZ$, the encoder takes a historical sequence of joint positions and velocities together with the residual torques between the nominal model and the measured torques. While the joint states provide information about the recent system trajectory, the residual torque provides information about the unknown payload.

In the human-exoskeleton setting, we instead aim to learn the full dynamics of the system. Both joint torque sensors and nominal models are usually unavailable. Therefore, we use pressure insole sensor data, namely the vertical ground reaction force $\ForceInsole$, and the two-dimensional center of pressure $\CopInsole \in \Real^2$. These measurements provide information about contact state and human inertia, which can be used to infer a latent representation.

\subsection{Learned Insole Mapping (LIM)}
Similar to legged robots, human movement involves intermittent contacts with the environment. For a single contact, the corresponding contact torques can be expressed as
\begin{equation}
    \label{eq:joint_total_contact}
    \jointTauContact = \JacContact\Transp{}(\jointPos)\ForceContact{}
\end{equation}
where $\JacContact$ denotes the contact Jacobian and $\ForceContact$ the contact force.

In practice, the contact force $\ForceContact$ is difficult to measure directly. For humans, it is typically obtained using force plates, which restrict measurements to laboratory environments. Instead, we use the measured vertical ground reaction force $\ForceInsole$ provided by the exoskeleton. Although $\ForceInsole$ is only a partial observation of the contact force, it provides information about the contact state and loading of the system. We therefore learn a mapping $\InsoleLearnedMap$ from this measurement to the corresponding contact torque, conditioned on the joint configuration $\jointPos$ and context $\latentVariableZ$:
\begin{equation}
    \label{eq:learned_insole_map}
    \jointTauContact = \InsoleLearnedMap(\jointPos, \latentVariableZ, \paramInsole)\ForceInsole\text{,}
\end{equation}
where $\paramInsole$ denotes the learnable parameters of the insole mapping, which are optimized jointly with the other model parameters.

Finally, Figure~\ref{fig:exolan_architecture} provides an overview of the ExoLaN architecture, combining the Context-Aware DeLaN with a TCN context encoder and the LIM.

\section{Experiments}
\label{sec:experiments}
In this section, we describe the dataset and training procedure, evaluate ExoLaN for human intention estimation and human trajectory estimation, and investigate the task information encoded in the learned latent representation.

\subsection{Dataset and Training Procedure}
We use the dataset provided by \cite{tcn_nature_2024}, which contains recordings from 21 users performing 28 tasks, sampled at 200~Hz. The tasks are categorized into three groups: cyclic, impedance-like, and unstructured movements. Ground-truth human joint torques are obtained using OpenSim~\cite{opensim} and are used as the target for training and evaluating the dynamics models.

The original study analyzes the informativeness of the different tasks and identifies seven tasks that provide approximately 95\% of the performance obtained using all 28 tasks. These seven tasks are walking, standing, calisthenics, push-pull, jump in place, turning, and cutting.
Following this selection, we use data from 14 users performing the seven selected tasks for training. The remaining seven users are held out for evaluation, providing a user-independent test set. We evaluate the model on the seven trained tasks and on the remaining 21 tasks for out-of-distribution evaluation.

Unless otherwise stated, the same user split and task split are used throughout the experiments, including human intention estimation, forward dynamics prediction, and task identification.

Following the approach from \cite{tcn_nature_2024} and to demonstrate the adaptation capability of ExoLaN, we combine both legs datasets to learn a single model, where $\latentVariableZ$ encodes task, user, and leg-specific information.

\subsection{Human Intention Estimation}
\label{subsec:exp_human_intention}
We first evaluate ExoLaN for human intention estimation. We consider the following methods:
\begin{itemize}
    \item ExoLaN (DeLaN-TCN, DeLaN-LSTM): context-aware + physics, where the LSTM-based variant follows \cite{cadelac}
    \item MLP-TCN, MLP-LSTM: context-aware + black-box
    \item TCN, LSTM: historical sequence augmented with the current state, following \cite{tcn_nature_2024}
    \item DeLaN: no history + physics, similar to \cite{pilan}
    \item MLP: no history + black-box mapping from $\genPos$, $\genVel$, $\genAcc$
\end{itemize}
For each method, we evaluate variants with and without LIM, yielding 16 methods in total. For the methods with DeLaN, we use the input transformation
$\mathbf{T}_\jointPos(\jointPos) = [\cos(\jointPos); \sin(\jointPos)]$. Both the inertia and potential energy networks in DeLaN use two hidden layers with 64 neurons per layer. The MLP baselines use two hidden layers with 96 neurons per layer, resulting in approximately the same number of trainable parameters. The LIM is implemented as an MLP with two hidden layers of 16 neurons per layer.
\begin{figure}[b]
    \includegraphics{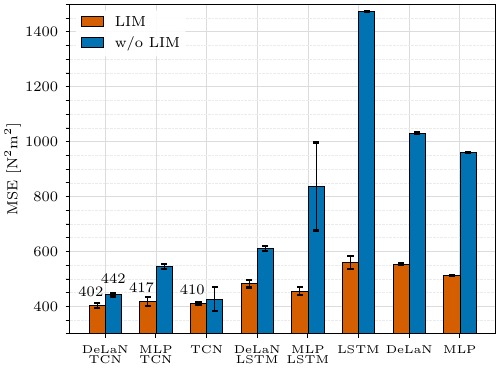}
    \caption{Torque estimation MSE on seven unseen users across the seven tasks seen during training, averaged over five random seeds, comparing models with and without LIM. DeLaN-TCN and DeLaN-LSTM, both with LIM, are two variants of ExoLaN.}
    \label{fig:inv_dyn_main7}
\end{figure}

\begin{figure}[t]
    \includegraphics{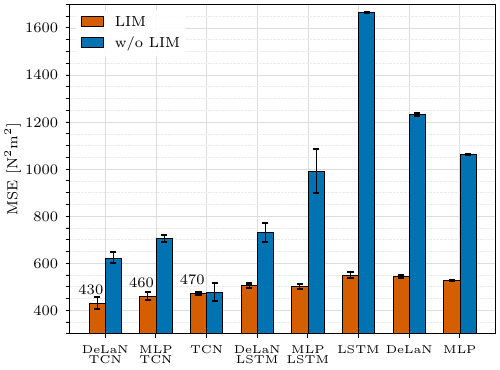}
    \caption{Torque estimation MSE on seven unseen users across 21 tasks not seen during training, averaged across the evaluation samples and five random training seeds, comparing models with and without LIM. DeLaN-TCN and DeLaN-LSTM, both with LIM, are two variants of ExoLaN.}
    \label{fig:inv_dyn_untrained_tasks21}
\end{figure}
\begin{figure*}[h]
    \centering
    \begin{minipage}{\textwidth}
        \centering
        \begin{subfigure}{0.30\textwidth}
            \includegraphics[width=\linewidth]{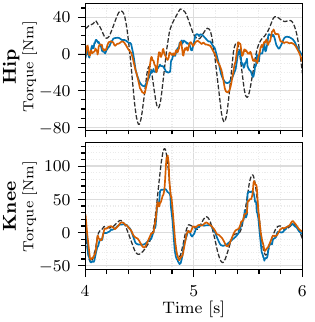}
            \caption{Calisthenics}
            \label{subfig:total_traj_calisthenics}
        \end{subfigure}
        \begin{subfigure}{0.22\textwidth}
            \includegraphics[width=\linewidth]{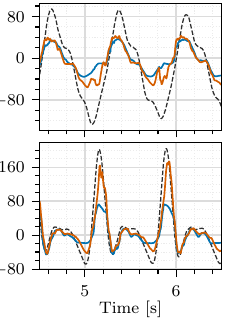}
            \caption{Run}
            \label{subfig:torque_traj_run}
        \end{subfigure}
        \begin{subfigure}{0.22\textwidth}
            \includegraphics[width=\linewidth]{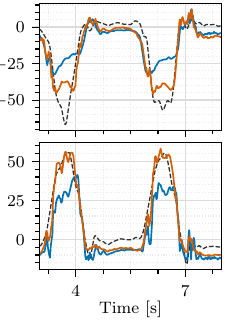}
            \caption{Squat}
            \label{subfig:torque_traj_squats}
        \end{subfigure}
        \begin{subfigure}{0.22\textwidth}
            \includegraphics[width=\linewidth]{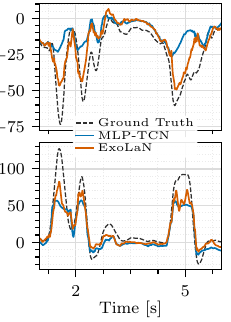}
            \caption{Ball Toss }
            \label{subfig:torque_traj_ball_toss}
        \end{subfigure}
    \end{minipage}
    \caption{Human intention estimation on unseen users across four tasks, using MLP-TCN and ExoLaN, both with LIM. The MSE values are reported as MLP-TCN vs.\ ExoLaN in $\text{N}^2\text{m}^2$: calisthenics ($1130$ vs.\ $\mathbf{1104}$), running ($4680$ vs.\ $\mathbf{2875}$), squats ($441$ vs.\ $\mathbf{146}$), and ball toss ($605$ vs.\ $\mathbf{300}$). Only calisthenics was included in the training set.}
    \label{fig:torque_traj_plots}
\end{figure*}

For training, all models are optimized for 1000 epochs to estimate the total actuated torque, i.e., $\jointTauHuman + \jointTauExo$. The human torque is then obtained by subtracting the exoskeleton torque $\jointTauExo$ from the estimated total torque. To account for differences in torque magnitude across users with different masses, we normalize the torque error by the user's mass before computing the squared error. The remaining training hyperparameters are reported in Tab.~\ref{tab:context_train_parameters} in Appendix~\ref{appendix:train_hyper}.

Figure~\ref{fig:inv_dyn_main7} presents the mean squared error (MSE) of the estimated human joint torques across both joints, evaluated on data from seven unseen users. The evaluation considers the same seven tasks used during training, with the reported MSE averaged across the evaluation samples and five random training seeds. Across the evaluated models, incorporating LIM consistently improves torque estimation, indicating that the insole measurements provide useful information about the contact dynamics. Both ExoLaN variants outperform their corresponding MLP baselines, suggesting the benefit of incorporating physical structure into the learned architecture.

Adding context conditioning improves the performance of both DeLaN and MLP methods, while the choice of context encoder also influences performance. The models with TCN generally outperform their LSTM-based counterparts, suggesting that the TCN provides a more effective representation of the temporal dependencies present in the observed system dynamics, consistent with prior comparisons of TCNs and LSTMs~\cite{original_tcn}. In particular, ExoLaN with DeLaN-TCN achieves the lowest MSE among the evaluated models, improving the MSE by 2\% relative to TCN with LIM and by 4\% relative to MLP-TCN with LIM.

The TCN only model also performs well and shows limited improvement from incorporating LIM. In cyclic tasks, where joint trajectories exhibit strong temporal regularity, recent joint state history can provide substantial information for torque estimation even without an explicit dynamical model. In these tasks, TCN achieves the lowest MSE of $464~\text{N}^2\text{m}^2$, closely followed by ExoLaN with the DeLaN-TCN architecture and LIM, with an MSE of $468~\text{N}^2\text{m}^2$. In contrast, for non-cyclic tasks, ExoLaN achieves the lowest MSE of $411~\text{N}^2\text{m}^2$, compared with $450~\text{N}^2\text{m}^2$ for TCN. These results demonstrate that the explicit dynamical structure learned by ExoLaN captures additional information that is not encoded by temporal history alone, particularly when temporal regularities are less pronounced.


While the differences between the models are relatively small for the trained tasks, the advantage of the physics-based formulation becomes more pronounced when evaluating unseen tasks. Figure~\ref{fig:inv_dyn_untrained_tasks21} presents the torque estimation MSE on the same unseen users performing 21 tasks that were not included during training. ExoLaN achieves a larger improvement over the MLP-TCN and TCN baselines, reducing the MSE by 7\% and 9\%, respectively. MLP-TCN also outperforms the TCN by 2\%, indicating improved generalization from combining nonlinear mapping with temporal context. These results demonstrate that the physical structure imposed by the DeLaN provides a useful inductive bias for generalization to tasks beyond those observed during training.

Figure~\ref{fig:torque_traj_plots} shows the estimated torque trajectories for four different tasks using ExoLaN and MLP-TCN with LIM. Among these tasks, only calisthenics, Fig.~\ref{subfig:total_traj_calisthenics}, was included in the training set. Overall, ExoLaN more accurately captures the torque amplitude, particularly for the knee joint in the unseen tasks, Fig.~\ref{subfig:torque_traj_run}-\ref{subfig:torque_traj_ball_toss}, while also providing reasonable estimates of the hip torque.

\subsection{Human Trajectory Estimation}
\label{subsec:exp_human_traj}
In this subsection, we evaluate the use of ExoLaN for predicting future system states, demonstrating its potential as a simulator.

In addition to the inverse dynamics loss used in Subsection~\ref{subsec:exp_human_intention}, we introduce a forward dynamics loss and a multi-step loss. The latter acts as a regularizer and improves long-horizon prediction~\cite{lutter2021learningdynamicsmodelsmodel}. Due to the additional computational cost of multi-step prediction, $\Subsamples$ rollout starting points are randomly sampled from the training set at each batch update. The resulting training objective is
\begin{align}
    \paramAll^* &= \arg \min_{\paramAll}
    \frac{1}{N}
    \sum_{i=1}^{N}\bigg[\left\| \genTau_i - \FID(\genPos_i, \genVel_i, \genAcc_i, \paramAll) \right\|^2_{\NormTorque}
    \label{eq:opt}\\
    &+ 
    \WeightFD
    \left\| \genAcc_i - \FFD(\genPos_i, \genVel_i, \genTau_i, \paramAll) \right\|^2\bigg]\notag
    \\
    &+  \frac{\WeightMS}{\Subsamples \MShorizon} \sum_{i=1}^{\Subsamples} \sum_{k=1}^{\MShorizon} \bigg[
    \| \genPos_{i+k} - \hat{\genPos}_{i+k} \|_{\NormQp}^{2}
    +
    \| \genVel_{i+k} - \hat{\genVel}_{i+k} \|_{\NormQv}^{2}
    \bigg]
    \notag
\end{align}
where $\paramAll = [\paramInertia, \paramPot, \paramLstm, \paramInsole]$, $\NormTorque$ denotes the user's mass normalization applied to the torque loss to balance the contribution of users with different body masses; $\NormQp$ and $\NormQv$ denote the diagonal covariance matrices of $\genPos$ and $\genVel$ in the training set, respectively. The weights $\WeightFD$ and $\WeightMS$ balance the contributions to the total loss from the forward dynamics and multi-step losses, respectively, and $\MShorizon$ denotes the multi-step prediction horizon. The predicted states $\hat{\genPos}_{i+k}$ and $\hat{\genVel}_{i+k}$ are obtained autoregressively by integrating the predicted accelerations over the $\MShorizon$ steps using semi-implicit Euler integration. 

The three ExoLaN variants are trained for 1000 epochs with a prediction horizon of $\MShorizon=4$ and $\Subsamples=256$.

\begin{figure}[t]
    \centering

    \begin{subfigure}{\columnwidth}
        \centering
        \includegraphics[width=\linewidth]{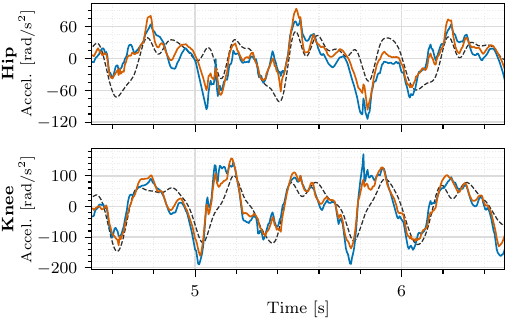}
        \caption{Run}
        \label{subfig:fwd_dyn_run}
    \end{subfigure}

    \vspace{0.5em}

    \begin{subfigure}{\columnwidth}
        \centering
        \includegraphics[width=\linewidth]{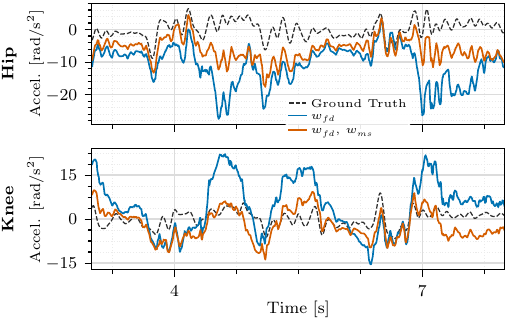}
        \caption{Squat}
        \label{subfig:fwd_dyn_squat}
    \end{subfigure}

    \caption{Human acceleration prediction for two tasks. MSE values are reported as MLP-TCN vs.\ ExoLaN in $\text{rad}^2\text{s}^{-4}$: running ($2841$ vs.\ $\mathbf{1394}$) and squats ($194$ vs.\ $\mathbf{68}$).}
    \label{fig:acc_traj_plots}
\end{figure}

Table~\ref{tab:fwd_dyn_loss_6config} reports the MSE of human torque and acceleration estimations, as well as position and velocity prediction at future steps $4$, $10$, and $20$.
ExoLaN trained only with the inverse dynamics loss achieves the lowest torque MSE. However, it performs poorly in forward dynamics, as indicated by the high acceleration and multi-step prediction errors. In principle, either the forward or inverse dynamics loss alone should suffice to obtain a model that performs well for both function approximations.
This mismatch can be attributed to two factors. First, we do not have ground-truth total torques $\jointTau$ and accelerations. The total torques are a composition of the ground truth human torque, exoskeleton torque, and the contact torque $\jointTauContact$ estimated via LIM. The accelerations are obtained by applying a Savitzky-Golay filter to the measured velocities. Consequently, the torque and acceleration targets are not guaranteed to be physically consistent with each other. 
Second, minimizing only the torque error does not explicitly constrain the learned inertia matrix to be well-conditioned for forward dynamics. 
The positive-definite inertia matrix is parameterized through a Cholesky factor $\inertiaTri$, with diagonal $\epsilon$ used for numerical stability. Although this guarantees positive definiteness, the resulting inertia matrix can still become poorly conditioned for some joint configurations, making the matrix inversion required for forward dynamics numerically sensitive and leading to inaccurate acceleration predictions.

By introducing the forward dynamics loss weighted by $\WeightFD$, ExoLaN substantially improves forward dynamics prediction, reducing the acceleration MSE by more than six orders of magnitude and substantially improving position and velocity rollout accuracy, at the cost of a relatively small $9\%$ increase in torque MSE.
\begin{table*}[t]
\centering
\scriptsize
\setlength{\tabcolsep}{4pt}
\caption{Forward dynamics evaluation on seven unseen users across all 28 tasks for ExoLaN trained with inverse dynamics only, with single-step forward dynamics loss $\WeightFD = 0.01$; and $\WeightFD = 0.01$ combined with multi-step loss $\WeightMS = 1$ ($\MShorizon = 4$).}
\label{tab:fwd_dyn_loss_6config}
\begin{tabular}{lcccccccc}
\toprule
\multirow{2}{*}{Model} & $\genTau$ & $\genAcc$ & \hspace{6pt}$\genPos_{4}$ & $\genVel_{4}$ & \hspace{6pt}$\genPos_{10}$ & $\genVel_{10}$ & \hspace{6pt}$\genPos_{20}$ & $\genVel_{20}$ \\ & {\scriptsize [$\text{N}^2\text{m}^2$]} & {\scriptsize [$\text{rad}^2\text{s}^{-4}$]} & \hspace{6pt}{\scriptsize [$\text{rad}^2$]} & {\scriptsize [$\text{rad}^2\text{s}^{-2}$]} & \hspace{6pt}{\scriptsize [$\text{rad}^2$]} & {\scriptsize [$\text{rad}^2\text{s}^{-2}$]} & \hspace{6pt}{\scriptsize [$\text{rad}^2$]} & {\scriptsize [$\text{rad}^2\text{s}^{-2}$]} \\
\midrule
ExoLaN & \textbf{434} & 4.97$\PosSciExp{8}$ & \hspace{6pt}2.78$\NegSciExp{1}$ & 2352 & \hspace{6pt}16.6 & 3.68$\PosSciExp{4}$ & \hspace{6pt}459 & 2.55$\PosSciExp{5}$ \\
ExoLaN ($\WeightFD$) & 475 & 337 & \hspace{6pt}1.41$\NegSciExp{3}$ & 0.184 & \hspace{6pt}7.63$\NegSciExp{3}$ & 2.42 & \hspace{6pt}0.0817 & 79.7 \\
ExoLaN ($\WeightFD$, $\WeightMS$) & 507 & \textbf{139} & \hspace{6pt}\textbf{1.38}$\mathbf{\NegSciExpB{3}}$ & \textbf{0.0787} & \hspace{6pt}\textbf{6.62}$\mathbf{\NegSciExpB{3}}$ & \textbf{0.900} & \hspace{6pt}\textbf{0.0329} & \textbf{5.79} \\
\bottomrule
\end{tabular}
\end{table*}


Finally, incorporating the multi-step loss through $\WeightMS$ further reduces the acceleration MSE by approximately 59\%. This improvement is also reflected qualitatively in Fig.~\ref{subfig:fwd_dyn_run}-\ref{subfig:fwd_dyn_squat}, which compares the predicted and ground-truth accelerations for two representative tasks, Run and Squat. For Run, the predictions from the models with and without $\WeightMS$ are relatively similar, although the multi-step loss provides a slightly closer match to the ground truth. The improvement is more significant for Squat, which was not included in the training set. In this case, incorporating $\WeightMS$ substantially improves the acceleration prediction, showing that the multi-step loss improves generalization to unseen tasks.

\begin{figure}[t]
    \centering
    \begin{subfigure}{\columnwidth}
        \centering
        \includegraphics[width=\linewidth]{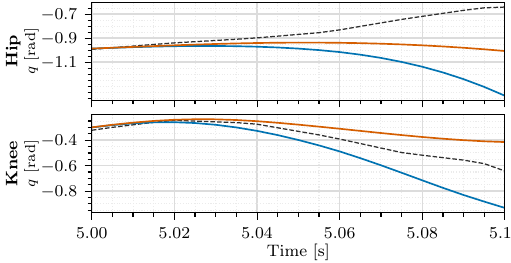}
        \caption{Joint position rollout}
        \label{subfig:fwd_dyn_open_loop_q}
    \end{subfigure}
    
    \par\vspace{1.0em}
    
    \begin{subfigure}{\columnwidth}
        \centering
        \includegraphics[width=\linewidth]{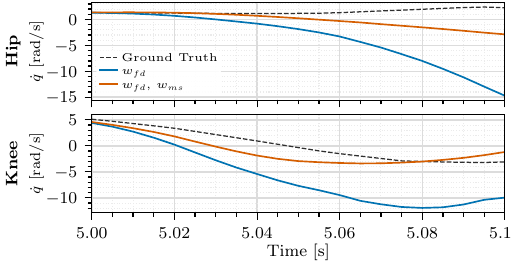}
        \caption{Joint velocity rollout}
        \label{subfig:fwd_dyn_open_loop_qdot}
    \end{subfigure}
    \caption{Open-loop autoregressive prediction of human joint states in the Run task for ExoLaN with only $\WeightFD$ and ExoLaN with both $\WeightFD$ and $\WeightMS$. The predicted joint positions and velocities are obtained by recursively integrating the estimated accelerations from an initial state, without further correction from measurements.}
    \label{fig:open_loop_plots}
\end{figure}

The multi-step loss also substantially improves the accuracy of the autoregressive rollouts. Compared with ExoLaN trained with only $\WeightFD$, the velocity MSE is reduced by 57\%, 63\%, and 93\% at future steps 4, 10, and 20, respectively, while the position MSE is reduced by 60\% at the 20-step horizon. Notably, the training loss uses a relatively short horizon of only four steps, yet the resulting model exhibits substantially improved prediction accuracy at the longer 10- and 20-step horizons. This indicates that the multi-step objective improves the stability of the learned dynamics beyond the horizon directly optimized during training. The improvement in long-horizon prediction is further illustrated in Fig.~\ref{fig:open_loop_plots}, which shows the position and velocity rollouts for the Run task. With the inclusion of $\WeightMS$, the predicted trajectories remain close to the ground truth for approximately 60~ms (12 steps), whereas with only $\WeightFD$, the predictions begin to diverge after approximately 30~ms (6 steps).

\subsection{Task Identification}
The learned latent representation is intended to capture information about the context of the human-exoskeleton system. To investigate whether this representation also contains information about the task being performed by the user, we evaluate its ability to distinguish between the different tasks in the dataset. Such task information could potentially be exploited to enable task-aware or adaptive assistive control, in which the assistance strategy and autonomy level could be adjusted according to the performed task.

For this analysis, we use the ExoLaN and MLP-TCN with LIM trained in Subsection~\ref{subsec:exp_human_intention}.For each trained model, we extract the corresponding inferred latent representations over the training trajectories and construct a dataset for task classification. Each latent dataset is split into 80\% for training and 20\% for testing. With five random seeds for each architecture, this results in 10 latent datasets in total. A separate task classifier is trained for each dataset, resulting in five classifiers for each architecture.

Each classifier is implemented as an MLP with three hidden layers of 64 neurons and is trained using a softmax cross-entropy loss for 1000 epochs. The remaining training hyperparameters are reported in Tab.~\ref{tab:classifier_train_parameters} in  Appendix~\ref{appendix:train_hyper}. The classifier is trained to identify the seven tasks considered in the training set. Since the number of samples is not uniform across tasks, class weighting is applied to the training loss to compensate for the resulting task imbalance.

Table~\ref{tab:classifier_summary_simple_task_id} reports the mean and standard deviation of the test accuracy across the five seeds for each architecture. Both latent representations contain significant task-related information, achieving test accuracies of 67.3\% for ExoLaN and 64.2\% for MLP-TCN. Both models substantially outperform the majority-class baseline of 31.0\%, corresponding to the most frequent task, jumping in place.
\begin{table}[h]
    \centering
    \caption{Task classification accuracy from the learned latent representations, averaged over five random seeds.}
    \begin{tabular}{lcc}
    \toprule
    Model & Majority Guess & Test Accuracy \\
    \midrule
    ExoLaN & \multirow{2}{*}{0.310} & \textbf{0.673$\pm$0.023} \\
    MLP-TCN &  & 0.642$\pm$0.021 \\
    \bottomrule
    \end{tabular}
    \label{tab:classifier_summary_simple_task_id}
\end{table}

The higher classification accuracy of ExoLaN suggests that incorporating the physical structure of the dynamics may promote more task-discriminative latent representations. These results support the use of the learned latent representation as a signal for task-adaptive assistive control.





\section{CONCLUSIONS}
\label{sec:conclusion}

We introduced ExoLaN, a Context-Aware DeLaN with a TCN-based context encoder and a Learned Insole Mapping (LIM) for modeling human-exoskeleton dynamics. We evaluated ExoLaN for torque estimation, corresponding to inverse dynamics, on seven unseen users across both tasks seen during training and previously unseen tasks. The results demonstrate improved generalization across users and tasks, while highlighting the benefit of incorporating physical structure and the LIM. We further showed that the learned dynamics can be used for forward dynamics prediction, providing a unified model for both inverse and forward dynamics.

As future work, we will integrate ExoLaN with computed-torque control for assistive exoskeletons. We will also investigate whole-body formulations, such as \cite{felan}, and explore ExoLaN as a physics-based simulator for imitation learning~\cite{human_gait_il} and reinforcement learning~\cite{leem2026exoplore}.



\section*{APPENDIX}

\subsection{Training Hyperparameters}
\label{appendix:train_hyper}
The training hyperparameters for human intention and trajectory estimation are reported in Tab.~\ref{tab:context_train_parameters}, while those for task classification are reported in Tab.~\ref{tab:classifier_train_parameters}.

\begin{table}[ht]
\centering
\caption{Training hyperparameters for human intention and trajectory training.}
\begin{tabular}{ll|ll}
\toprule
\textbf{Parameter} & \textbf{Value} & \textbf{Parameter} & \textbf{Value} \\
\midrule
Activation & Tanh & TCN Kernel Size & 2 \\
Batch Size & 8192 & LSTM Hidden Layers & $16 \times 5$ \\
Learning Rate & $1 \times 10^{-3}$ & History Length & 63 \\
Weight Decay & $1 \times 10^{-3}$ & Context Dim. ($\latentVariableZ$) & 16 \\
TCN Hidden Layers & $16 \times 5$ & Training Samples & 1.4M \\
\bottomrule
\end{tabular}
\label{tab:context_train_parameters}
\end{table}

\begin{table}[ht]
\centering
\caption{Training hyperparameters for task classification using inferred latent representations.}
\begin{tabular}{ll|ll}
\toprule
\textbf{Parameter} & \textbf{Value} & \textbf{Parameter} & \textbf{Value} \\
\midrule
Activation & ReLU & Epochs & 1000 \\
Batch Size & 4096 & Hidden Layers & $64 \times 3$ \\
Learning Rate & $1 \times 10^{-3}$ & Class Weighting & Inv.\ Train Freq. \\
Weight Decay & $1 \times 10^{-4}$ & Training Samples & 1.1M \\
\bottomrule
\end{tabular}
\label{tab:classifier_train_parameters}
\end{table}

\addtolength{\textheight}{-1cm}   







\addtolength{\textheight}{-1.0cm}
\bibliographystyle{./bibtex/IEEEtran}
\bibliography{references}

\end{document}